\documentclass[10pt,twocolumn,letterpaper]{article}

\usepackage{iccv}
\usepackage{times}
\usepackage{epsfig}
\usepackage{graphicx}
\usepackage{amsmath}
\usepackage{amssymb}
\usepackage{multirow}
\usepackage{diagbox}

\usepackage[breaklinks=true,bookmarks=false]{hyperref}

\iccvfinalcopy 

\def\iccvPaperID{****} 

\ificcvfinal\fi

\begin{document}

\title{Prior Aided Streaming Network for Multi-task Affective Recognition\\ at the 2nd ABAW2 Competition}

\author{Wei Zhang\textsuperscript{\rm 1}\footnotemark[1] , 
Zunhu Guo\textsuperscript{\rm 1,2}\footnotemark[1] , 
Keyu Chen\textsuperscript{\rm 1}\footnotemark[1]\thanks{Equal Contribution.}, 
Lincheng Li\textsuperscript{\rm 1}, 
Zhimeng Zhang\textsuperscript{\rm 1}, 
Yu Ding\textsuperscript{\rm 1}\footnotemark[2]\thanks{Corresponding Author.}\\
 \textsuperscript{\rm 1}Virtual Human Group, Netease Fuxi AI Lab\\
 \textsuperscript{\rm 2}Southwest University, China \\
{\tt\small \{zhangwei05,guozunhu,chenkeyu02,lilincheng,zhangzhimeng,dingyu01\}@corp.netease.com}\\

}

\maketitle
\ificcvfinal\thispagestyle{empty}\fi

\begin{abstract}
   Automatic affective recognition has been an important research topic in human computer interaction (HCI) area. With recent development of deep learning techniques and large scale in-the-wild annotated datasets, the facial emotion analysis is now aimed at challenges in the real world settings. In this paper, we introduce our submission to the 2nd Affective Behavior Analysis in-the-wild (ABAW2) Competition. In dealing with different emotion representations, including Categorical Emotions (CE), Action Units (AU), and Valence Arousal (VA), we propose a multi-task streaming network by a heuristic that the three representations are intrinsically associated with each other. Besides, we leverage an advanced facial expression embedding as prior knowledge, which is capable of capturing identity-invariant expression features while preserving the expression similarities, to aid the down-streaming recognition tasks. The extensive quantitative evaluations as well as ablation studies on the Aff-Wild2 dataset prove the effectiveness of our proposed prior aided streaming network approach.
\end{abstract}

\section{Introduction}

Recognizing and analyzing facial affective statements from human behaviors is a long-standing problem in the intersection area of the computer science and psychology community. An ideal human-computer interaction system is expected to capture the vivid human emotions, mostly conveyed by facial performances, and to react respectively. Because of the diverse environments and varying contexts where emotions occur, the perception of facial effectiveness is always natural to our human beings but never straightforward to the artificial intelligent machines. Thanks to the continuous research of psychology and rapid development of deep learning methods, especially recent published large scale in-the-wild annotated datasets e.g., \textit{Aff-Wild}~\cite{zafeiriou2017aff} and \textit{Aff-Wild2}~\cite{kollias2019expression}, the automatic affective recognition approaches are now pushed to meet the real-world requirements. 

Different from most existed facial emotion datasets~\cite{jaffe, zhang2014bp4d,zhang2016multimodal,jiang2020dfew} that contain only one of the three common used emotional representations: Categorical Emotions (CE), Action Units (AU), and Valence Arousal (VA), the \textit{Aff-Wild2}~\cite{kollias2019expression} dataset is annotated with all three kinds of emotional labels, containing extended facial behaviors in random conditions and increased subjects/frames to the former \textit{Aff-Wild}~\cite{zafeiriou2017aff} dataset. Consequently, the multi-task affective recognition can benefit from it, for example, the works~\cite{deng2020multitask,gera2020affect,zhang2020m,saito2020action} participated in the first Affective Behavior Analysis in-the-wild (ABAW) Competition~\cite{kollias2020analysing}.

In this work, we propose a novel multi-task affect recognition framework for the ABAW2 Competition~\cite{2106.15318}. In contrast to the previous methods which take the multiple emotion recognition problems as parallel tasks, we design our algorithm pipeline in a streaming structure to fully exploit the hierarchical relationships among the three representation spaces. Specifically, we make our single-flow network first estimates the action units from input images, then the emotion labels, and finally the VA distribution. Such arrangements are made due to a heuristic that the regressing order AU$\to$CE$\to$VA should match the underlying semantic level of three target emotion representations. For instance, the facial action coding system (FACS) defines AU based on local patches and therefore AU-related features could provide low-level information for the global categorical emotion (CE) classification task. Moreover, the seven-dimensional emotion distributions (spanned by the categorical classes) can be compressed into 2D with the two principle components: Valence and Arousal (VA). 

Another contribution of our framework is that we utilize an advanced facial expression embedding model to employ helpful prior knowledge for the downstream tasks, i.e., AU detection, CE classification, and VA regression. Despite the traditional facial expression recognition (FER) models have regressed continuous expression distributions for discrete classification, they can hardly encode the fine-grained expression features. In this work, we adopt the triplet-based expression embedding~\cite{zhang2021learning} model as the backbone of the entire framework. Since the expression embedding is trained to distinguish minor expression similarities between different subjects, it can provide powerful expression-related priors to the high-level emotion recognition task. 

In participating the second ABAW2 Competition, we conduct extensive experiments on the \textit{Aff-Wild2}\cite{kollias2019expression} dataset. In order to improve the generalization ability of our multi-task model, we augment the training dataset with BP4D~\cite{zhang2014bp4d}, BP4D+~\cite{zhang2016multimodal}, DFEW~\cite{jiang2020dfew} and AffectNet~\cite{mollahosseini2017affectnet}. Because of the multi-task framework and streaming design, each module of our network can be fine-tuned on images with no need for all the three emotion representation labels to exist. 

In sum, the contributions of this work are two-folds:
\begin{itemize}
    \item We propose a streaming network to handle the multi-task affect recognition problem. By heuristically designing the regression order, the streaming structure allows to exploit inner relationships across different emotional spaces.
    \item We employ an identity-invariant expression prior model as backbone. With fine-grained expression related features, our network can well capture the high-level information for the emotional recognition tasks.
\end{itemize}

\section{Related Works}
In this section, we briefly review some concepts, works and datasets related with the affective recognition problem.   
\subsection{Emotional Representation}
Human affective behavior analysis has attracted great interest in Human-Computer Interaction. With the help of effective emotional representation, the computer will gain a better understanding of how human brain behave, leading to the user friendly experience between humans and machines. There are three common used emotional representations: 7 basic emotion categories~\cite{ekman2003darwin}, Facial Action Units (AUs)~\cite{FACS} and 2-D Valence
and Arousal (VA) Space~\cite{russell1980circumplex}. The 7 basic emotions includes Anger, Disgust, Fear, Happiness, Sadness, Surprise and Neutral. AUs~\cite{FACS} include 32 atomic facial action descriptors based on facial muscle groups, which facilitate the physical and fine-grained understanding of human facial expressions. The detection of facial AU occurrence offers crucial information for emotion recognition~\cite{AU4EmoctionRecog}, micro-expression detection~\cite{AU4MicroExpression}, and mental health diagnosis~\cite{AUD4diagnose}. The Valance in VA space represents the degree of emotional positiveness and negativeness and the Arousal shows whether the emotion is passive or active. 
\subsection{Affect Annotation Dataset}
The 2nd Workshop and Competition on Affective Behavior Analysis in-the-wild (ABAW)~\cite{zafeiriou2017aff,kollias2019expression, kollias2020analysing, kollias2019face, kollias2019deep, kollias2021affect, kollias2021distribution} provides a benchmark dataset Aff-Wild2 for three recognition challenges: 7 basic emotion classification, 12 AUs detection and VA regression. Extened from Aff-wild~\cite{zafeiriou2017aff}, Aff-wild2 increases the number of annotated videos with 545 videos annotated by valence-arousal, 539 videos annotated by 7 basic emotion categories and 534 videos annotated by 12 AUs. Aff-Wild2 is currently the largest in the wild dataset in aspect of VA, AU and basic emotion expressions.  
\subsection{Automatic Affective Behavior Analysis}
The challenges of affective behavior analysis has attracted lots of research efforts. We will briefly introduce some related works. Kuhnke~\etal~\cite{kuhnke2020two} use the multi-model information of vision and audio, proposing a two-stream aural-visual network for multi-task training. Considering the problems of unbalanced data and missing label, Deng~\etal~\cite{deng2020multitask} propose a structure of Teacher-Student to learn from the unlabelled data by way of soft label. Besides the multi-task frameworks, Gera~\etal~\cite{gera2020affect} focus on the task of discrete emotion classification and propose the network based on attention mechanism.  Zhang~\etal~\cite{zhang2020m} propose a multi-model approach $M^{3}T$ for valence-arousal estimation using the visual feature extracted from 3D convolution network and a bidirectional recurrent neural network and the audio features extracted from a acoustic sub-network. Saito~\etal~\cite{saito2020action} tackle the problem of AUs label inconsistency, proposing a Pseudo-intensity Model to learn the degree of facial appearance change and a mapping model to predict the AUs. 

\section{Method}

\begin{figure*}
    \centering
    \includegraphics[width=1\linewidth]{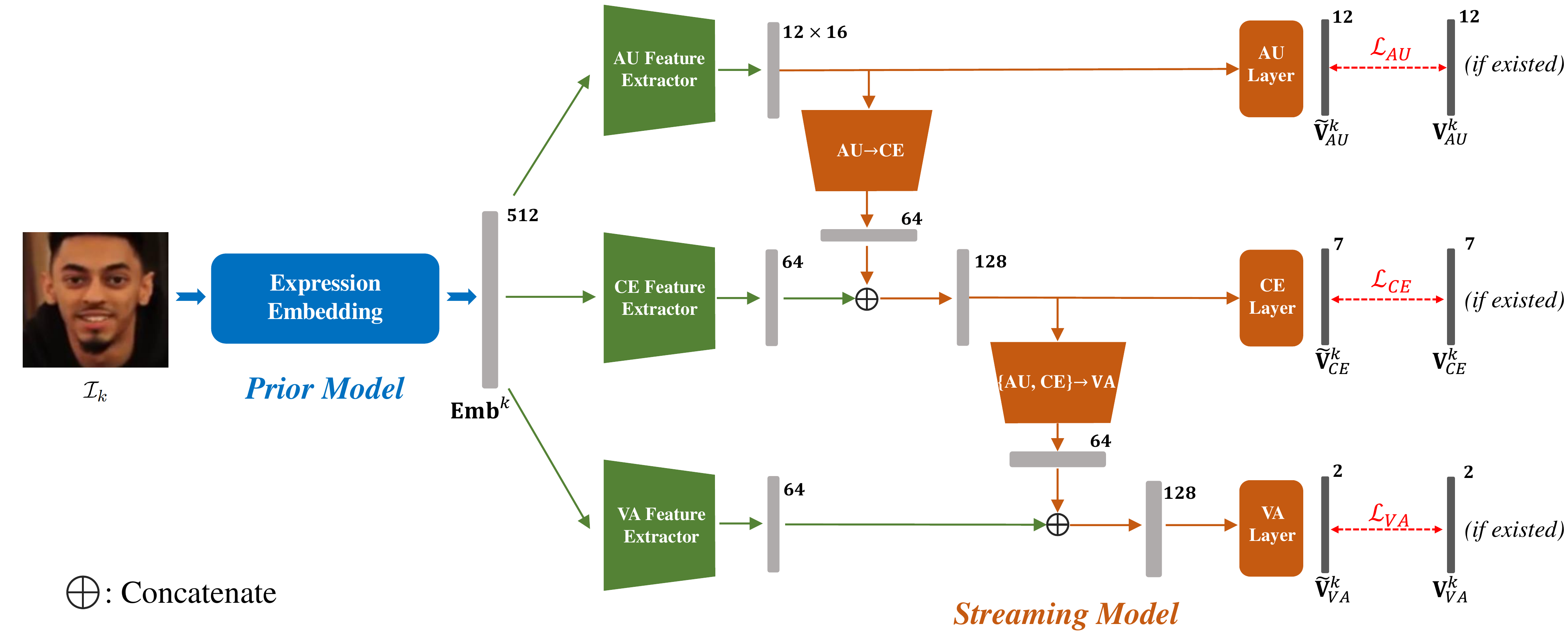}
    \caption{To be updated.}
    \label{fig:pipeline}
\end{figure*}

In this section, we introduce our method for affective behavior analysis in the 2nd ABAW2 Competition. The overall pipeline is illustrated in Fig.~\ref{fig:pipeline}. The entire framework consists of two components: a prior model for extracting prior expression embedding knowledge and a streaming model for exploiting the hierarchical relationships among three emotional representations.

\subsection{Overview}

As described in the official white paper~\cite{2106.15318}, the ABAW2 Competition contains three challenges, corresponding to the three commonly used emotion representations: seven categorical emotions, twelve action units, and two dimensional valence arousal. We propose a general framework to jointly handle the three individual tasks. Despite the different psychological philosophies of the three emotional representations, it is widely agreed that the representations are intrinsically associated with each other~\cite{tian2001recognizing}. One of the evidence is that the similar facial muscle movements (action units) mostly indicate the similar inner statements, and so the perceived facial emotions. However, most previous research works on multi-task emotion recognition omit this fact and they just model the different tasks in parallel branches. Inspired by the observation above, we design the recognition process in a serial manner AU$\to$CE$\to$VA, from local action units to global emotion states. The streaming structure is helpful to adjust the hierarchical distributions on different feature levels. For example, the optimizing energy from the most high-level VA space should be back-propagated to the low-level features and thus help the other two tasks in training.

Due to the limited subjects and unbalanced annotations of existed affective datasets, it is a challenging issue to prevent the emotion recognition model from overfitting on the disturbing factors, like background or random noise. To tackle this problem, we adopt a prior facial expression embedding model~\cite{zhang2021learning}, which can capture the detailed expression similarities across different people, into our framework. The expression embedding brings at least two advantages. First, by training on even larger facial image datasets with the identity invariant constraint, the embedding itself is independent to the identity attributes and therefore can improve the network's generalizability to unseen subjects. Second, the expression embedding model~\cite{zhang2021learning} is targeted for discriminating the minor expression similarities within triplet training data. It provides a nice initialization for our latter emotion recognition tasks.

Combining with the prior and the streaming model, we train our multi-task affective recognition model in an end-to-end manner. Given an image $\mathbf{\mathcal{I}}_k$ with at least one of the three emotional annotations $\{\mathbf{V}_{AU}^k, \mathbf{V}_{CE}^k, \mathbf{V}_{VA}^k\}$, we send it to the full network for training and compute corresponding losses on its existed labels. In the following, we will introduce the network structure and loss functions in detail.

\subsection{Prior Model}

We adopt the Deviation Learning Network (DLN) from~\cite{zhang2021learning} as the expression prior to our framework. In order to generate a compact and continuous expression embedding space disentangled from the identity factor, the DLN model are trained on more than 500 thousands of annotated triplets from the FECNet~\cite{vemulapalli2019compact} dataset. 

Following the idea from~\cite{vemulapalli2019compact,zhang2020facial}, the DLN aims to map the similar expression image pair (\textit{anchor} and \textit{positive}) close to each other in the low-dimensional space, while keep the dissimilar expression image pair (\textit{anchor} and \textit{negative}) away from each other. To efficiently exclude the identity attributes from the extracted image features, the DLN model proposes a deviation module by subtracting the identity vectors (produced by a pre-trained face recognition model) from the facial ones. 

Since the original DLN model maps the facial expression images into a 16 dimensional space, which leaves quite tight room for optimization in our problem, we only take the pre-trained deviation module from~\cite{zhang2021learning} that produces 512 dimensional features. Specifically, given a facial image $\mathbf{\mathcal{I}}_k$ from training dataset, the prior model is expected to generate a 512 dimensional embedding vector $\mathbf{Emb}^k$ that contains identity-invariant expression information. In training the entire framework, we also make the expression embedding model to be trainable and adaptively adjust the embedding vector results.

\subsection{Streaming Model}

With prior generated expression embedding vector, we first construct three individual feature extractor to downsample $\mathbf{Emb}^k$ from 512 to $12\times 16$, $64$, $64$, respectively. 

We start from the AU branch and introduce our streaming regression process on each of three tasks. For AU features in $\mathbb{R}^{12\times 16}$, we directly send it into a multilayer perceptron (MLP) to regress for the AU score per each of twelve classes. Denote the final output of AU predictions as $\mathbf{\tilde{V}}_{AU}^k = \{\tilde{v}_1, \tilde{v}_1,..., \tilde{v}_{12}\} \in \mathbb{R}^{12}$ and the ground-truth AU label $\mathbf{V}_{AU}^k=\{v_0, v_1,...,v_12\}\in \{0, 1\}^{12}$, we apply the multi-label cross entropy loss~\cite{he2018joint} as following:

\begin{equation}
\begin{split}
    \mathcal{L}_{AU}=&\log(1+\sum_{i\in \Omega_{0}}e^{\tilde{v}_i})+\log(1+\sum_{j\in \Omega_{1}}e^{-\tilde{v}_j}),\\
    \text{where}&\\
    \Omega_0 =& \{~i~~ |~~ \text{if}~~ v_i=0~\},\\
    \Omega_1 =& \{~j~~ |~~ \text{if}~~ v_j=1~\}.
\end{split}
\end{equation}

On the other hand, the AU features are sent into the CE branch after translated by the AU$\to$CE model. We concatenate the translated AU features ($\mathbb{R}^{64}$) with CE ones ($\mathbb{R}^{64}$) to be a joint vector. Then the 128 dimensional features are sent into CE layers for emotion classification. The output CE possibility vector $\mathbf{\tilde{V}}_{CE}^k$ and the annotated emotion label $\mathbf{V}_{CE}^k$ is evaluated by the Softmax Classifier Loss:

\begin{equation}
    \mathcal{L}_{CE}=\text{Softmax}(\mathbf{\tilde{V}}_{CE}^k, \mathbf{V}_{CE}^k).
\end{equation}

The \{AU,CE\}$\to$VA model takes the CE joint features as input and generates another 64 dimensional feature to aid the VA regression. Similar to the last operation on CE task, we concatenate the VA features with the translated ones and send them into the VA layers. Defining the two dimensional vector output as $\mathbf{\tilde{V}}_{VA}^k=\{\tilde{v},\tilde{a}\}$ and the ground-truth as $\mathbf{V}_{VA}^k=\{v,a\}$, the VA loss is computed by the Concordance Correlation Coefficient (CCC) metric:

\begin{equation}
    \mathcal{L}_{VA}=CCC_v + CCC_a.
\end{equation}

The total loss of the streaming network can be formulated as:

\begin{equation}
    \mathcal{L}_{total} = \alpha_{AU} \cdot \mathcal{L}_{AU} + \alpha_{CE} \cdot \mathcal{L}_{CE} + \alpha_{VA} \cdot \mathcal{L}_{VA},
\end{equation}

where $\alpha_{\cdot}$ is boolean valueable indicating the existence of groundtruth label on each track.

\subsection{Algorithm Details}

\noindent \textbf{Data Augmentation}. In addition to the original training set of \textit{Aff-Wild2}~\cite{kollias2019expression}, our model is further trained on the BP4D~\cite{zhang2014bp4d}, BP4D+~\cite{zhang2016multimodal}, DFEW~\cite{jiang2020dfew}, and AffectNet~\cite{mollahosseini2017affectnet}. While processing the external datasets, we only keep the annotated classes that are consistent with the \textit{Aff-Wild2}~\cite{kollias2019expression}.

\noindent \textbf{Pseudo Label}. Another approach we proposed for alleviating the overfitting / data unbalancing issue is to generate reliable pseudo labels for training. We exploit the underlying relationships between AU and CE. Particularly, some AUs are always mapped to the same CE. In this way, we can quickly infer the missing CE labels from explicit AU annotations.

\section{Experiments}

\begin{table}[t]
\begin{center}
\begin{tabular}{c|c|c|c} \hline
\textbf{Method} & \textbf{AU}  & \textbf{CE} & \textbf{VA}\\ \hline
Baseline~\cite{2106.15318} & 0.310 & 0.366 & 0.220 \\
Ours w/o prior  & 0.464 & 0.718 & 0.422  \\
Ours w/o streaming & 0.677 & 0.677 & 0.447  \\
Ours & \textbf{0.742} & \textbf{0.790} & \textbf{0.495}  \\
\hline\\
\end{tabular}
\caption{Ablation comparison to our method w/o prior model or streaming structure and the baseline~\cite{2106.15318}. The best result per each track is indicated in bold. }
\label{tab:ablation}
\end{center}
\end{table}

\begin{table*}[t]
\begin{center}
\begin{tabular}{c|ccc|ccc|ccc} \hline
\multirow{2}*{\diagbox{\textbf{Validation set}}{\textbf{Track}}} &\multicolumn{3}{c|}{\textbf{AU}} & \multicolumn{3}{c|}{\textbf{CE}} &  \multicolumn{3}{c}{\textbf{VA}}\\
~ & F1 & \textit{TAcc} & Score & F1 & \textit{TAcc} & Score & $CCC_V$ & $CCC_A$ & Score  \\ \hline
Original & 0.588 & 0.896 & 0.742 & 0.757 & 0.856 & 0.790 & 0.488 & 0.502 & 0.495 \\
Fold-1 & - & - & 0.753 & - & - & \textbf{0.783} & - & - & 0.578 \\
Fold-2 & - & - & \textbf{0.772} & - & - & 0.725 & - & - & 0.591 \\
Fold-3 & - & - & 0.755 & - & - & 0.762 & - & - & 0.532 \\
Fold-4 & - & - & 0.753 & - & - & 0.770 & - & - & \textbf{0.621} \\
Fold-5 & - & - & 0.758 & - & - & 0.765 & - & - & 0.606 \\

\hline\\
\end{tabular}
\caption{Quantitative results of our prior aided streaming on different validation sets. The best result per each track is indicated in bold.}
\label{tab:validation}
\end{center}
\end{table*}

In this section, we give some experimental results based on the validation dataset of \textit{Aff-Wild2}~\cite{kollias2019expression}, as well as the 5-fold cross validation results. As part of submission to the ABAW2 Competition, we also upload our code for open release.

\subsection{Training}

We processed all videos in the \textit{Aff-Wild2} dataset into frames by OpenCV and employ the OpenFace~\cite{baltrusaitis2018openface} detector to eatract and resize all facial images into $224\times 224$ scale. We trained the entire framework on a NVIDIA RTX 3090 graphics card for around 20 hours.

\subsection{Results}

There are two kinds of validation set to be evaluated in our experiment. One is the official provided validation set, the other is the 5-fold cross validation set. We report both quantitative results in Tab.~\ref{tab:validation}. 

In order to evaluate the effectiveness of our proposed algorithm design, i.e., prior model and streaming network, we conduct ablation studies by comparing the models trained without the components. The quantitative results shown in Tab.~\ref{tab:ablation} indicate that both modules help to improve the recognition/classification performance on each emotion representation track.

\section{Conclusion}

In this paper, we introduce our deep learning based framework for multi-task affective recognition in the second ABAW2 Competition. We propose a streaming network by exploiting the hierarchical relationships between different emotion representations. Besides, we employ an expression prior model to improve the generalization ability of our model to the test set. The quantitative comparisons prove that each component is effective to the affective recognition tasks. We have also presented the experimental results on the official validation dataset.

{\small
\bibliographystyle{unsrt}
\bibliography{egbib}
}

\end{document}